\def\year{2019}\relax
\documentclass[letterpaper]{article} 
\usepackage{aaai19}  
\usepackage{times}  
\usepackage{helvet}  
\usepackage{courier}  

\usepackage{booktabs} 
\usepackage{url}
\usepackage{graphicx}  
\usepackage{amsfonts} 
\usepackage{amsmath}
\usepackage{bm}
\usepackage{multirow}
\usepackage{subfigure}
\usepackage{algorithm}  
\usepackage{algpseudocode}  
\usepackage{makecell}
\usepackage{xspace}
\usepackage{amssymb}
\usepackage{array}
\usepackage{footmisc}
\usepackage{listings}
\usepackage[np, autolanguage]{numprint}
\usepackage{paralist}
\usepackage{enumitem}
\usepackage{colortbl}
\usepackage{soul}
\usepackage{epstopdf}
\usepackage{pifont}
\usepackage{xcolor}
\usepackage{mathrsfs}
\usepackage{todonotes}
\usepackage{subfig}
\usepackage{subfloat}

\newcommand{\aka}{\emph{a.k.a.,}\xspace}

\newcommand{\etal}{\emph{et al.}\xspace}
\newcommand{\ignore}[1]{}

\newcommand{\dubbelop}{$^{\blacktriangle}$}

\newcommand{\citet}[1]
{\citeauthor{#1}~\shortcite{#1}}
\newcommand{\citep}{\cite}

\begin{document}
    %
    \title{Abstractive Text Summarization by Incorporating Reader Comments}
    \author{Shen Gao,\textsuperscript{1}
    	Xiuying Chen,\textsuperscript{1}
    	Piji Li,\textsuperscript{3}
    	Zhaochun Ren,\textsuperscript{4}
    	Lidong Bing,\textsuperscript{5}
    	Dongyan Zhao,\textsuperscript{1,2}
    	Rui Yan\thanks{Corresponding author: Rui Yan (ruiyan@pku.edu.cn)}\textsuperscript{1,2}\\
    	\textsuperscript{1}{Institute of Computer Science and Technology, Peking University, Beijing, China} \\
    	\textsuperscript{2}{Center for Data Science, Peking University, Beijing, China}\\
    	\textsuperscript{3}{Tencent AI Lab, Shenzhen, China} \\
    	\textsuperscript{4}{JD.com, Beijing, China} \\
    	\textsuperscript{5}{R\&D Center Singapore, Machine Intelligence Technology, Alibaba DAMO Academy}\\
    	\{shengao, xy-chen, zhaody, ruiyan\}@pku.edu.cn,
    	pijili@tencent.com, 
    	renzhaochun@jd.com,
    	l.bing@alibaba-inc.com}
            
    \maketitle
    \begin{abstract}
    In neural abstractive summarization field, conventional sequence-to-sequence based models often suffer from summarizing the wrong aspect of the document with respect to the main aspect.
    To tackle this problem, we propose the task of reader-aware abstractive summary generation, which utilizes the reader comments to help the model produce better summary about the main aspect. 
    Unlike traditional abstractive summarization task, reader-aware summarization confronts two main challenges: (1) Comments are informal and noisy;
    (2) jointly modeling the news document and the reader comments is challenging.
    To tackle the above challenges, we design an adversarial learning model named reader-aware summary generator (RASG), which consists of four components: (1) a sequence-to-sequence based summary generator; (2) a reader attention module capturing the reader focused aspects; (3) a supervisor modeling the semantic gap between the generated summary and reader focused aspects; (4) a goal tracker producing the goal for each generation step.
    The supervisor and the goal tacker are used to guide the training of our framework in an adversarial manner.
    Extensive experiments are conducted on our large-scale real-world text summarization dataset, and the results show that RASG achieves the state-of-the-art performance in terms of both automatic metrics and human evaluations. The experimental results also demonstrate the effectiveness of each module in our framework.
    We release our large-scale dataset for further research\footnote{\url{http://t.cn/EAH5JxS}}.
    \end{abstract}

    \section{Introduction}

Abstractive summarization can be regarded as a sequence mapping task that the source text is mapped to the target summary, and has drawn much attention since the deep neural networks are widely applied in natural language processing field.
Recently, sequence-to-sequence (seq2seq) framework~\cite{Sutskever2014SequenceTS} has been proved effective for the task of abstractive summarization~\cite{Chopra2016AbstractiveSS,see2017get} and other text generation tasks~\cite{Tao2018GetTP,Gao2019Product}.
In this paper, we use ``\textbf{aspect}'' to denote the topic described in a specific paragraph or a sentence of a news document, and use ``\textbf{main aspect}'' to denote the central topic which the author tends to convey to the readers.
Although a document may describe an event in many different aspects, the summary of this document should always focus on the main aspect.
As shown in Table~\ref{tab:intro-case}, the good summary describes the main aspect and the bad summary describes another trivial aspect that is not the main point of the document.
To focus on the main aspect, some summarization methods~\cite{Sun2018AUM,Zhou2017SelectiveEF,Bansal2018FastAS} first select several sentences about the main aspect and then generate the summary.
However, it is very challenging to discover which is the main aspect of the news document.

\begin{table}[t]
\centering
\caption{Examples of the text summarization. The text in red denotes the focused aspect by the good summary, while the text in blue is described by the bad summary. The text with underline is the focused aspect by reader comments.} 
\scriptsize
\begin{tabular}{l|l}
\toprule
document & \multicolumn{1}{p{6cm}}{
On August 28, according to a person familiar with the matter, \textcolor{red}{\ul{Toyota Motor Corporation will invest 500 million U.S. dollars into the Uber}, a taxi service company, with a \ul{valuation of up to 72 billion U.S. dollars.}} 
The investment will focus on driverless car technology.
However, its development path is not smooth.
In March of this year, \textcolor{blue}{a Uber driverless car hit a woman and caused her death.}
In last year, Softbank also invested into Uber with a valuation of \$48 billion.
}     \\ 
\hline
\multicolumn{1}{l|}{\multirow{2}{*}{comments}} & \multicolumn{1}{p{6cm}}{Toyota's investment in Uber is a wise choice.}     \\ \cline{2-2} 
\multicolumn{1}{l|}{}                         & \multicolumn{1}{p{6cm}}{\$500 million investment is really a lot of money!}      \\ \hline
\textcolor{red}{good} summary                                     & \multicolumn{1}{p{6cm}}{Toyota invests \$500 million into Uber with a valuation of \$72 billion}   \\ \hline 
\textcolor{blue}{bad} summary                                     & \multicolumn{1}{p{6cm}}{An Uber driverless car hits a passerby to death}   \\
\bottomrule
\end{tabular}
\label{tab:intro-case}
\end{table}

Nowadays, a great number of news comments are generated by readers to express their opinions about the event. 
Some comments may mention the main aspect of the document for several times. 
Take the case in Table~\ref{tab:intro-case} as an example, the focused aspect of the reader is ``investment of Toyota'' which is also the main aspect of this document. 
To be specific, we define ``\textbf{reader focused aspect}'' to denote the focused aspect by a reader through the comments.
Intuitively, these reader comments may help the summary generator capture the main aspect of document, thereby improving the quality of the generated summary. 
Therefore, in this paper, we investigate a new problem setting of the task of abstractive text summarization.
We name such paradigm of extension as \emph{reader-aware abstractive text summarization}.


The effect of comments or social contexts in document summarization have been explored by several previous works~\cite{Hu2008CommentsorientedDS,Yang2011SocialCS,Li2015ReaderAwareMS,li2017reader}.
Unlike these approaches that directly extract sentences from the original document~\cite{Hu2008CommentsorientedDS,Yang2011SocialCS,Li2015ReaderAwareMS}, we aim to generate a natural-sounding summary from scratch instead of extracting words from the document.

Generally, existing text summarization approaches confront two challenges when addressing reader-aware summarization task.
The first challenge is that reader comments are very noisy and informative. 
Not all the information provided by the comments is useful when modeling the reader focused aspects. 
Therefore, it is crucial to make the model own the ability of capturing main aspect and filtering noisy information when incorporating reader comments. 
The second challenge is how to generate summaries by jointly modeling the main aspect of document and the reader focused aspect revealed by comments.
Meanwhile, the model should not be sensitive to the diverse unimportant aspects introduced by some reader comments. 
Thus, simply absorbing all the reader aspect information to directly guide the model to generate summary is not feasible, as it will make the generator lose the ability of modeling the main aspect.

In this paper, we propose a summarization framework named \emph{reader-aware summary generator} (RASG) that incorporates reader comments to improve the summarization performance.
Specifically, a seq2seq architecture with attention mechanism is employed as the basic summary generator.
We first calculate alignment between the reader comments words and document words, and this alignment information is regarded as reader attention representing the ``\textbf{reader focused aspect}''.
Then, we treat the decoder attention weights as the focused aspect of the generated summary, \aka ``\textbf{decoder focused aspect}''.
After each decoding step, a supervisor is designed to measure the distance between the reader focused aspect and the decoder focused aspect.
Given this distance, a goal tracker provides the goal to the decoder to induce it to reduce this distance.
The training of our framework RASG is conducted in an adversarial way.
To evaluate the performance of our model, we collect a large amount of document-summary pairs associated with several reader comments from social media website. 
Extensive experiments conducted on this dataset show that RASG significantly outperforms the state-of-the-art baselines in terms of ROUGE metrics and human evaluations.
 
To sum up, our contributions can be summarized as follows: 

$\bullet$ We propose a reader-aware abstractive text summarization task. To solve this task, we propose an end-to-end learning framework to conduct the reader attention modeling and reader-aware summary generation.

$\bullet$ 
We design a supervisor as well as a goal tracker to guide the generator to focus on the main aspect of the document.

$\bullet$ To reduce the noisy information introduced by the reader comments, we propose a denoising module to identify which comments are helpful for summary generation automatically.

$\bullet$ We release a large scale abstractive text summarization dataset associated with reader comments.
Experimental results on this dataset demonstrate the effectiveness of our proposed framework.
    
\section{Related Work}
Text summarization can be classified into extractive and abstractive methods. 
Extractive methods~\cite{Jadhav2018ExtractiveSW,Narayan2018RankingSF} read the article and get the representations of the sentences and article to select sentences.
However, summaries generated by extractive methods always suffer from redundancy problem.
Recently, with the emergence of neural network models for text generation, a vast majority of the literature on summarization is dedicated to abstractive summarization~\cite{Bansal2018FastAS,Ma2018AutoencoderAA,Zhou2018SequentialCN}.
On the text summarization benchmark dataset CNN/DailyMail, the state-of-the-art abstractive methods outperform the best extractive method in terms of ROUGE score.
Most methods for abstractive text summarization are based on the sequence-to-sequence model~\cite{Sutskever2014SequenceTS}, which encodes the source texts into the semantic representation with an encoder, and generates the summaries from the representation with a decoder.
To tackle the out-of-vocabulary problem, some researchers employ the copy mechanism to copy some words from the input document to summary~\cite{Gu2016IncorporatingCM,see2017get}.
To capture the main aspect of document, Chen~\etal~\shortcite{Bansal2018FastAS} propose to select salient sentences and then rewrite these sentences to a concise summary.
This approach achieves the state-of-the-art of text summarization on CNN/DailyMail benchmark dataset.
Unlike document summarization that needs to encode a long text, social media summarization usually reads short and noisy text and has become a popular task these days.
After Hu~\etal~\shortcite{Hu2015LCSTSAL} propose a short text summarization dataset on social media and many researchers follow this task.
Lin~\etal~\shortcite{Lin2018GlobalEF} propose a seq2seq based model which uses an CNN to refine the representation of source context.
Wang~\etal~\shortcite{Wang2018ART} use convolutional seq2seq model to summarize text and use the policy gradient algorithm to directly optimize the ROUGE score.
However, these summarization models do not utilize the reader's comments in generating summaries.

To consider the reader's comments into text summarization, the reader-aware summarization is proposed and it mainly takes the form of extractive approaches. 
Graph-based method has been used for comment oriented summarization task such as ~\cite{Hu2007CommentsorientedBS,Hu2008CommentsorientedDS},  where they identify three relations (topic, quotation, and mention) by which comments can be linked to one another.
Recently, Nguyen~\etal~\shortcite{Nguyen2016SoLSCSumAL} publish a small extractive sentence-comment dataset which can not be used to train neural models due to its small size. 
Li~\etal~\shortcite{Li2015ReaderAwareMS} propose an unsupervised compressive multi-document summarization model using sparse coding method.
Following previous work, there are some models~\cite{li2017reader,Li2017SalienceEV} using variational auto-encoder to model the latent semantic of original article and reader comments.
Different from our abstractive summarization task, these related works are all based on extractive or compressive approaches.

\section{Problem Formulation}
\label{sec:formulation}

Before presenting our approach for the reader-aware summarization, we first introduce our notations and key concepts. 


To begin with, for a document $X^d=\{x^d_1, x^d_2, \dots, x^d_{T^d}\}$, we assume there is a comment set $X^c=\{c_1, c_2, \dots, c_{T^c}\}$ where $c_i = \{x^c_{i,1}, x^c_{i,2}, \dots, x^c_{i,T^c_{i}}\}$ is the $i$-th comment, $x^d_i$ denotes the $i$-th word in document $X^d$, and $x^c_{i,j}$ denotes the $j$-th word in $i$-th comment sentence $c_i$.
Given the document $X^d$, the summary generator reads the comments $X^c$, then generates a summary $\hat{Y} = \{\hat{y}_1, \hat{y}_2, \dots, \hat{y}_{T^{\hat{Y}}}\}$.
Finally, we use the difference between generated summary $\hat{Y}$ and ground truth summary $Y$ as the training signal to optimize the model parameters.

\section{The Proposed RASG Model}

\subsection{Overview}

\begin{figure*}
    \centering
    \includegraphics[scale=0.66]{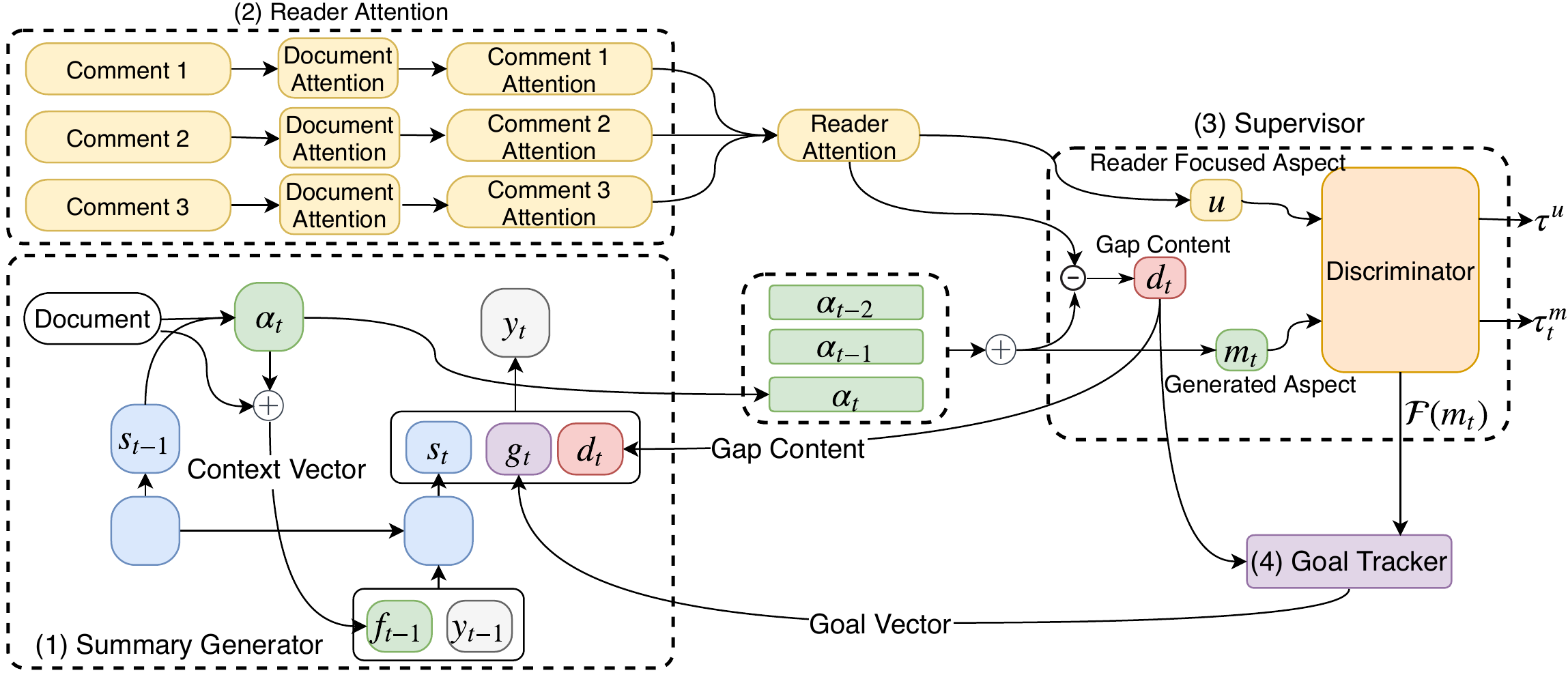}
    \caption{Overview of RASG. We divide our model into four parts: (1) \textit{Summary generator} generates a summary to describe the main aspect of document. (2) \textit{Reader attention} module models the readers attention of document. (3) \textit{Supervisor} models the gap of focused document aspect between generated summary and reader comments. (4) \textit{Goal tracker} sets a goal of summary generator according to gap given by supervisor. 
    }
    \label{fig:overview}
\end{figure*}

In this section, we propose our \emph{reader-aware summary generator}, abbreviated as RASG. 
The overview of RASG is shown in Figure~\ref{fig:overview} which can be split into four main parts:

$\bullet$ \textit{Summary generator} is a seq2seq based architecture with attention and copy mechanisms.

$\bullet$ \textit{Reader attention} module learns a semantic alignment between each word in document and comments, thus captures the reader focused aspect.

$\bullet$ \textit{Supervisor} measures the semantic gap between decoder focused aspect and reader focused aspect.
There is also a discriminator which uses convolutional neural network to extract features and then distinguishes how similar is decoder focused aspect to reader focused aspect.

$\bullet$ \textit{Goal tracker} utilizes the semantic gap learned by supervisor and the features extracted learned by the discriminator to set a goal, which is further utilized as a more specific guidance for summary generator to produce better summary.


\subsection{Summary generator}

At the beginning, we use an embedding matrix $e$ to map one-hot representation of each word in the document $X^d$ and comments $X^c$ to a high-dimensional vector space.
We denote $e(x)$ as the embedding representation of word $x$.
From these embedding representations, we employ a bi-directional recurrent neural network (Bi-RNN) to model the temporal interactions between words:
\begin{align}
    h^d_t &= \text{Bi-RNN}_d(e(x^d_t), h^d_{t-1}) ,
\end{align}
\noindent where $h^{d}_t$ denotes the hidden state of $t$-th step in Bi-RNN for document $X^d$. 
We denote the final hidden state $h^d_{T^d}$ of $\text{Bi-RNN}_d$ as the vector representation of the document $X^d$.
Following~\cite{see2017get,Ma2018AHE}, we choose the long short-term memory (LSTM) as the Bi-RNN cell. 

Then we apply a linear transform layer on the input document vector representation $h^d_{T^d}$ and use the output of this layer as the initial state of decoder LSTM, shown in Equation~\ref{equ:dec-init}.
In order to reduce the burden of compressing document information into initial state $s_0$, we use the attention mechanism~\cite{Bahdanau2014NeuralMT} to summarize the input document into context vector $f_{t-1}$ dynamically and we will show the detail of these in the following sections.
We then concatenate the context vector $f_{t-1}$ with the embedding of previous step output $e(y_{t-1})$ and feed this into decoder LSTM, shown in Equation~\ref{equ:dec-step}.
We use the notion $\left[ \cdot ; \cdot \right]$ as the concatenation of two vectors.
\begin{align}
    s_0 &= W_d h^d_{T^d} + b_d, \label{equ:dec-init}\\
    s_t &= \text{LSTM} \left( s_{t-1}, [f_{t-1}; e(y_{t-1})] \right) . \label{equ:dec-step}
\end{align}
At $t$-th decoding step, we use the decoder state $s_{t-1}$ to attend to each the document states $h^d$ and resulting in the attention distribution $\alpha_{t} \in \mathbb{R}^{T^d}$, shown in Equation~\ref{equ:attention-sm}.
Then we use the attention distribution $\alpha_{t}$ to weighted sum the document states as the context vector $f_{t-1}$.
\begin{align}
    \alpha^{'}_{t, i} &= W_a^\intercal \tanh \left( W_s s_t + W_h h^d_i \right), \\ 
    \alpha_{t, i} &= \exp \left( \alpha^{'}_{t, i} \right) / \textstyle \sum^{T^d}_{j=1} \exp \left(\alpha^{'}_{t, j} \right), \label{equ:attention-sm}\\
    f_{t-1} &= \textstyle \sum_{i=1}^{T^d} \alpha_{t, i} h^d_i .
\end{align}
Finally, an output projection layer is applied to get the final generating distribution $P_{v}$ over vocabulary, as shown in Equation~\ref{equ:out-proj}.
We concatenate goal vector $g_{t}$, gap content $d_{t}$, and the output of decoder LSTM $s_t$ as the input of the output projection layer.
The goal vector $g_{t}$ represents the goal of current generation step, the gap content $d_{t}$ denotes the semantic gap between generated summary and reader focused document and we will show the details of these variables in the following sections.
\begin{equation}
    P_{v} = \text{softmax} \left( W_v [s_t; g_{t}; d_{t}]  + b_v \right), \label{equ:out-proj}
\end{equation}
In order to handle the out-of-vocabulary (OOV) problem, we equip the pointer network~\cite{Gu2016IncorporatingCM,vinyals2015pointer,see2017get} with our decoder, which makes our decoder capable to copy words from the source text.
The design of the pointer network is the same as the model used in~\cite{see2017get}, thus we omit this procedure in our paper due to the limited space.
We use the negative log-likelihood as the loss function:
\begin{align}
    \mathcal{L}_g &= - \textstyle \sum^{T^{\hat{Y}}}_{t=1} \log P_{v}(y_t).
\end{align}

\subsection{Denoising module}

Due to the fact that reader comments are a kind of informal text, they may consist of many noisy information, and not all the comments are helpful for generating better summaries.
Consequently, we employ a \textbf{denoising module} to distinguish which comments are helpful.
First, we employ an $\text{Bi-RNN}_c$ to model the comment word embeddings:
\begin{equation}
    h^c_{i,t} = \text{Bi-RNN}_c(h^c_{i,t-1}, e(x^c_{i,t})),
\end{equation}
where $h^c_{i,t}$ denotes the hidden state of $t$-th word in $i$-th comment $c_i$.
Next, we use average-pooling operation over these hidden states to produce a vector representation $a_i$ of $i$-th comment, shown in Equation~\ref{equ:comment-avg}.
Finally, we apply a linear transform with sigmoid function to predict whether the comment is useful, and the sigmoid output $\hat{\beta}_i \in (0, 1)$ also can be seen as a salience score of $i$-th comment given the document representation $h^d_{T^d}$.
\begin{align}
    a_i &= \text{avg} (\{ h^c_{i,1}, h^c_{i,2}, \dots, h^c_{i,T^c_i}\}), \label{equ:comment-avg} \\
    \hat{\beta}_i &= \text{sigmoid} (W_s [a_i; h^d_{T^d}] + b_i) , \label{equ:slience-score}
\end{align}
To train the denoising module, we use the cross entropy loss to supervise this procedure.
\begin{equation}
    \mathcal{L}_d = - \textstyle \sum^{T^c}_{t=1} \hat{\beta}_i \log (\beta_i). \label{equ:denoising-loss}
\end{equation} 
where $\beta_i \in \{0, 1\}$ is the ground truth salience score of comments. 
$\beta_i = 1$ denotes the $i$-th comment $c_i$ is helpful for generating summary and vice-versa. 

\subsection{Reader attention modeling}
To model the reader focused aspect, we first calculate the word alignment of reader comments towards the document.
We use the embeddings of words in document and comments to calculate the semantic alignment score. Precisely, $\gamma_{i, j, k}$ is the alignment socre between the $i$-th document word $x^d_i$ and the $k$-th word in the $j$-th comment $x^c_{j, k}$, as shown in Equation~\ref{equ:word-sim}:
\begin{align}
    \gamma_{i, j, k} &= e(x^d_i)^\top e(x^c_{j, k}), \label{equ:word-sim} \\ 
    \delta_{i, j} &= \max(\{\gamma_{i, j, 1}, \dots, \gamma_{i, j, T^c_j}\}), \label{equ:word-sim-max}
\end{align}
In Equation~\ref{equ:word-sim-max}, we use a max-operation over the alignment $\gamma_{i, j, \cdot}$ to signify whether the $i$-th word of document is focused by the $j$-th comment.
We regard the alignment score $\delta_{i, j}$ as the reader attention weight for the $j$-th reader comment to the $i$-th document word.

In order to reduce the interference caused by the noisy comments, we employ the comment salience score $\hat{\beta}_j$ obtained from the denoising module to weighted combine the $j$-th reader attention $\delta_{i, j}$, as shown in Equation~\ref{equ:word-sim-weight}. It means that noisy comments will contribute less in the procedure of reader attention modeling.
\begin{align}
    \epsilon^{'}_{i} &= \textstyle \sum^{T^c}_{j=1} \delta_{i, j} \hat{\beta}_j, \label{equ:word-sim-weight}\\
    \epsilon_{i} &= \exp \left( \epsilon^{'}_{i} \right) / \textstyle \sum^{T^d}_{j=1} \exp \left(\epsilon^{'}_{j} \right). \label{equ:word-sim-sm}
\end{align}
Finally, we get the reader attention $\epsilon_{i} \in \mathbb{R}$ for $i$-th document word after a softmax function as shown in Equation~\ref{equ:word-sim-sm}.

\subsection{Supervisor} \label{sec:supervisor}

To model the semantic gap between the generated summary and the reader focused aspects, we design a supervisor module.
First, for the decoder, we need to know which aspect in document has been focused by our summary generator in the past decoding steps.
We sum up the latest $k$ attention distributions $\{\alpha_t, \alpha_{t-1}, \dots, \alpha_{t-k+1}\}$ and result in $\nu_t \in \mathbb{R}^{T^d}$ as the focus distribution of generated summary over $T^d$ document words, shown in Equation~\ref{equ:current_attention}.
Then we use $\nu_t$ to weighted sum the document hidden states $h^d_{\cdot}$ and result in $m_t$:
\begin{align}
    \nu_t &= 1/k \textstyle \sum^{k-1}_{i=0} \alpha_{t-i} , \label{equ:current_attention}\\
    m_t &= \textstyle \sum^{T^d}_{i=1} \nu_{t,i} h^d_{i} ,
\end{align}
where $m_t$ represents the focused aspect by the latest $k$ decoding steps, \aka decoder focused aspect.

Next, we use the reader attention $\epsilon_{\cdot}$ to weighted sum the document hidden states $h^d_{\cdot}$:
\begin{equation}
    u = \textstyle \sum^{T^d}_{i=1} \epsilon_{i} h^d_i , \label{equ:reader-attention-weighted-doc}
\end{equation}
where $u$ represents the reader focused aspect.

For encouraging the decoder focused aspect become similar to the reader focused aspect, we employ an CNN based discriminator to signify the difference between the decoder focused aspect $m_t$ and the reader focused aspect $u$.
Then we can use this difference to guide the decoder focus on the reader focused aspect.
Typically, the discriminator is a binary classifier which can be decomposed into a convolutional feature extractor $\mathcal{F}$ shown in Equation~\ref{equ:feature-extractor} and a sigmoid classification layer shown in Equation~\ref{equ:discriminator-sigmoid-m} and \ref{equ:discriminator-sigmoid-u}. 
\begin{align}
    \mathcal{F}(x) &= \text{relu} (W_c \otimes x) , \label{equ:feature-extractor}\\ 
    \tau^m_t &= \text{sigmoid} (W_f \mathcal{F}(m_t) + b_f) , \label{equ:discriminator-sigmoid-m}\\
    \tau^u &= \text{sigmoid} (W_f \mathcal{F}(u) + b_f) , \label{equ:discriminator-sigmoid-u}
\end{align}
where $\otimes$ denotes the convolutional operation, trainable parameter $W_c$ denotes the convolutional kernel, and $\tau^m_t$ and $\tau^u$ are both the classification probabilities.

Note that a token generated at time $t$ will influence not only the gradient received at that time but also the gradient at subsequent time steps.
Intuitively, the decoding attention $\nu_t$ of latter decoding step is more similar to the attention of final summary than the earlier steps.
Thus we propose to define the cumulative loss with a discount factor $\varphi \in (0,1]$ as the loss functions.
Note that the training objective for discriminator can be interpreted as maximizing the log-likelihood for classification, whether the input $x$ in Equation~\ref{equ:feature-extractor} comes from reader focused aspect or from decoder focused aspect. 
\begin{align}
    \mathcal{L}^d_c &= \textstyle \sum^{T^d}_{t=1} \varphi^{T^d-t} (\log \tau^{u} + \log (1 - \tau^m_t)), \label{equ:dis-d}\\
    \mathcal{L}^g_c &= \textstyle \sum^{T^d}_{t=1} \varphi^{T^d-t} (\log (1 - \tau^m_t)). \label{equ:dis-c}
\end{align}

In order to model the gap between reader focused aspect and decoder focused aspect, we subtract the reader attention $\epsilon \in \mathbb{R}^{T^d}$ by $\nu_{t} \in \mathbb{R}^{T^d}$ resulting in attention difference $\zeta_t \in\mathbb{R}^{T^d}$, shown in Equation~\ref{equ:attention-gap}.
Then we use the attention difference $\zeta_t$ to sum up the document hidden states $h^d_{\cdot}$:
\begin{equation}
    \zeta_t = \epsilon - \nu_{t}, \label{equ:attention-gap} \quad
    d_t = \textstyle \sum^{T^d}_{i=1} \zeta_{t, i} h^d_i .
\end{equation}
where $d_t$ denotes the semantic of unfocused document aspects by summary generator, \aka gap content.
To encourage the summary generator focus on the unfocused document aspects, we feed the gap content $d_t$ to the generator, as shown in Equation~\ref{equ:out-proj}.

\subsection{Goal tracker} \label{sec:goal-tracker}

Since the discriminator only provides a scalar guiding signal $\tau^m_t$ at each decoding step, it becomes relatively less informative when the sentence length $T^{\hat{Y}}$ goes larger. 
Inspired by LeakGAN~\cite{Guo2018LongTG}, the proposed RASG framework allows discriminator to provide additional information, denoted as goal vector $g_t$.
In view of there is certain relationship between the goal of current decoding step and previous steps, we need to model the temporal interactions between the goal of each step.
More specifically, we introduce a \textbf{goal tracker} module, an LSTM that takes the extracted feature vector $\mathcal{F}(m_t)$ and gap content $d_t$ as its input at each step $t$ and outputs a goal vector $g_t$:
\begin{align}
    g_t &= \text{LSTM} (g_{t-1}, [\mathcal{F}(m_t); d_t]) .
\end{align}
In order to achieve higher consistency of reader focused aspect, we feed the goal vector $g_t$ into the generator to guide the generation of the next word, as shown in Equation~\ref{equ:out-proj}.

\subsection{Model training}

As our model is trained in an adversarial manner, we re-split the parameters in our model into two parts: (1) \textit{generation module} including the parameters of summary generator, reader attention module and goal tracker; (2) \textit{discriminator module} including the parameters of CNN classifier.
As for training generation module, we sum up the loss function of denoising module $\mathcal{L}_d$, cross entropy between ground truth $\mathcal{L}_g$ and the result of discriminator $\mathcal{L}^g_c$, as shown in Equation~\ref{equ:loss-generator}.
We use the $\mathcal{L}$ to optimize the parameters of generation module.
\begin{equation}
    \mathcal{L} = \mathcal{L}_g + \mathcal{L}_d + \mathcal{L}^g_c , \label{equ:loss-generator}
\end{equation}
Next, we train the discriminator module to maximize the probability of assigning the correct label to both generated aspect $m_t$ and reader focused aspect $u$.
More specifically, we optimize the parameters of discriminator module according to the loss function $\mathcal{L}^d_c$ calculated in Equation~\ref{equ:dis-d}.

\section{Experimental Setup}

\subsection{Research questions}

We list four research questions that guide the experiments: 
\noindent \textbf{RQ1}: Does RASG outperform other baselines?
\noindent \textbf{RQ2}: What is the effect of each module in RASG? 
\noindent \textbf{RQ3}: Does RASG capture useful information from noisy comments?
\noindent \textbf{RQ4}: Can goal tracker give a helpful guidance to decoder?

\subsection{Dataset}

We collect the document-summary-comments pair data from Weibo which is the largest social network website in China, and users can read a document and post a comment about the document on this website.
Each sample of data contains a document, a summary and several reader comments.
Most comments are about the readers' opinion of their focused aspect in the document.
In order to train the denoising module, we should give a ground truth label $\beta_i$ for $i$-th comment.
When there is at least one common word in summary and comment, we regard such comment is helpful for generating summary.
Accordingly, we give the $\beta_i = 1$ to $i$-th comment when it contains at least one common word and give $\beta_i = 0$ when it does not.
In total, our training dataset contains 863826 training samples.
The average length of document is 67.08 words, average length of comment is 16.61 words and average length of summary is 16.56 words.
The average comments number of a document is 9.11.

\subsection{Evaluation metrics}

For evaluation metrics, we adopt ROUGE score~\cite{lin2004rouge} which is widely applied for summarization evaluation~\cite{Sun2018AUM,chen2018iterative}. 
The ROUGE metrics compare generated summary with the reference summary by computing overlapping lexical units, including ROUGE-1 (unigram), ROUGE-2 (bi-gram) and ROUGE-L (longest common subsequence). 

\subsection{Comparison methods}

In order to prove the effectiveness of each module in RASG, we conduct some ablation models introduced in Table~\ref{tab:ablations}.

\begin{table}[t]
\centering
\caption{Ablation models for comparison.}
\label{tab:ablations}
\small
\begin{tabular}{@{}l@{~}l}
\toprule
Acronym & Gloss \\
\midrule
RASG w/o DM &  \multicolumn{1}{p{5cm}}{\small RASG w/o \textbf{D}enoising \textbf{M}odule}\\
RASG w/o G &  \multicolumn{1}{p{5cm}}{\small RASG w/o \textbf{G}ap content}\\
RASG w/o GT &  \multicolumn{1}{p{5cm}}{\small RASG w/o \textbf{G}oal \textbf{T}racker}\\
RASG w/o GTD &  \multicolumn{1}{p{5cm}}{\footnotesize RASG w/o \textbf{G}oal \textbf{T}racker \textbf{D}iscriminator}\\
\bottomrule
\end{tabular}
\end{table}

To evaluate the performance of our proposed dataset and model, we compare it with the following baselines:

\noindent (1) \textbf{S2S}: Sequence-to-sequence framework~\cite{Sutskever2014SequenceTS} has been proposed for language generation task. 
\noindent (2) \textbf{S2SR}: We simply add the reader attention $\epsilon_{\cdot}$ on attention distribution $\alpha_{t, \cdot}$ in each decoding step.
\noindent (3) \textbf{CGU}: Lin~\etal~\shortcite{Lin2018GlobalEF} propose to use the convolutional gated unit to refine the source representation, which achieves the state-of-the-art performance on social media text summarization dataset.
\noindent (4) \textbf{LEAD1}: LEAD1 is a commonly used baseline~\cite{Nallapati2017SummaRuNNerAR,see2017get}, which selects the first sentence of document as the summary.
\noindent (5) \textbf{TextRank}: Mihalcea~\etal~\shortcite{Mihalcea2004TextRankBO} propose to build a graph, then add each sentence as a vertex and use link to represent semantic similarity. 
Sentences are sorted based on final scores and a greedy algorithm is employed to select summary sentences.

\subsection{Implementation details}

We implement our experiments in TensorFlow~\cite{abadi2016tensorflow} on an NVIDIA P40 GPU. 
The word embedding dimension is set to 256 and the number of hidden units is 512.
We set the $k=5$ in the Equation~\ref{equ:current_attention} and $\varphi=0.5$ in Equation~\ref{equ:dis-d} and \ref{equ:dis-c}.
We use Adagrad optimizer~\cite{Duchi2010AdaptiveSM} as our optimizing algorithm.
We employ beam search with beam size 5 to generate more fluency summary sentence.

\section{Experimental Results}

\subsection{Overall performance}

\newcommand{\cbkgrnd}{\cellcolor{blue!15}}
\newcommand{\phantomtriangle}{\phantom{\dubbelop}}
\begin{table}[t]
\centering
\small
\caption{ROUGE scores comparison between baselines.}
\begin{tabular}{@{}l ccc @{}}
\toprule
& ROUGE-1 & ROUGE-2 & ROUGE-L \\
\midrule
S2S &  23.86 & 9.86 & 23.83 \\
S2SR &  24.70 & 10.00 & 24.50 \\
CGU & 27.32 & 11.36 & 25.49  \\
RASG & \textbf{30.33} & \textbf{12.39} & \textbf{27.16} \\
\midrule
LEAD1 & 5.51 & 1.71 & 4.94 \\
TextRank & 13.5 & 4.55 & 11.46 \\
\bottomrule
\end{tabular}
\label{tab:comp_rouge_baselines}
\end{table}

For research question \textbf{RQ1}, we examine the performance of our model in terms of ROUGE. 
Table~\ref{tab:comp_rouge_baselines} lists performances of all comparisons in terms of ROUGE score.
We see that RASG achieves a 11.0\%, 9.1\% and 6.6\% increment over the state-of-the-art method CGU in terms of ROUGE-1, ROUGE-2 and ROUGE-L respectively.
It is worth noticing that the baseline model S2SR achieves better performance than S2S which demonstrates the effectiveness of incorporating reader focused aspect in summary generation.
However when compared with RASG, S2SR achieves lower performance in terms of all ROUGE score.
Thus, simply adding the reader focused aspect into generation procedure is not a good reader-aware summarization method.

\subsection{Ablation study}

\begin{table}[t]
\centering
\small
\caption{ROUGE scores of different ablation models.}
\begin{tabular}{@{}lcc c@{}}
\toprule
& ROUGE-1 & ROUGE-2 & ROUGE-L \\
\midrule
RASG w/o DM & 27.29 & 11.01 & 24.64 \\
RASG w/o G & 28.03 & 11.24 & 25.28 \\
RASG w/o GT & 22.75 & 9.17 & 20.80 \\
RASG w/o GTD & 19.30 & 6.95 & 17.70 \\
RASG & \textbf{30.33} & \textbf{12.39} & \textbf{27.16} \\
\bottomrule
\end{tabular}
\label{tab:comp_rouge_ablation}
\end{table}
Next, we turn to research question \textbf{RQ2}.
We conduct ablation tests on the usage of denoising module, supervisor as well as the goal tracker and the ROUGE score result is shown in Table~\ref{tab:comp_rouge_ablation}.
The discriminator provides the scalar training signal $\mathcal{L}_c^g$ for generator training and the feature vector $\mathcal{F}(m_t)$ for goal tracker.
Consequently, there is an increment of 17.51\% from RASG w/o GTD to RASG w/o GT in terms of ROUGE-L, which demonstrates the effectiveness of discriminator.
As for the effectiveness of goal tracker, compared with RASG and RASG w/o GT, RASG w/o GTD offers a decrease of 45.23\% and 17.88\% in terms of ROUGE-1, respectively. This demonstrates that the goal tracker with the feature from discriminator plays an important role in producing better summary.
However, using the goal tracker without the feature extracted by the discriminator does not help improve the performance of the summary generator, shown by the performance of RASG w/o GTD.
Finally, RASG w/o DM offers a decrease of 10.22\% compared with RASG in terms of ROUGE-L, which demonstrates the effectiveness of denoising module. 

\subsection{Denoising ability}

Next, we turn to research question \textbf{RQ3}.
Due to the fact that the denoising module is learned in a supervised way, there is a ground truth label associated with each comment.
Thus when the predict salience score $\hat{\beta}_i > 0.5$ we classify it as a helpful comment and vice-versa.
As the denoising module can be regarded as a binary classifier to classify each comment to $\hat{\beta}_i = 1$ or $\hat{\beta}_i = 0$, we calculate the classification recall score of comments to measure the performance of this module.
The recall curve is shown in Figure~\ref{fig:subfig}.
As the training progresses, the recall score is on a steady upward curve which proves the improved performance of denoising module.
To conclude, the denoising module can give a meaningful salience score for the subsequent process.

\subsection{Analysis of goal tracker}


\begin{figure} 
  \centering 
  \subfigure[]{ 
    \includegraphics[width=0.45\linewidth]{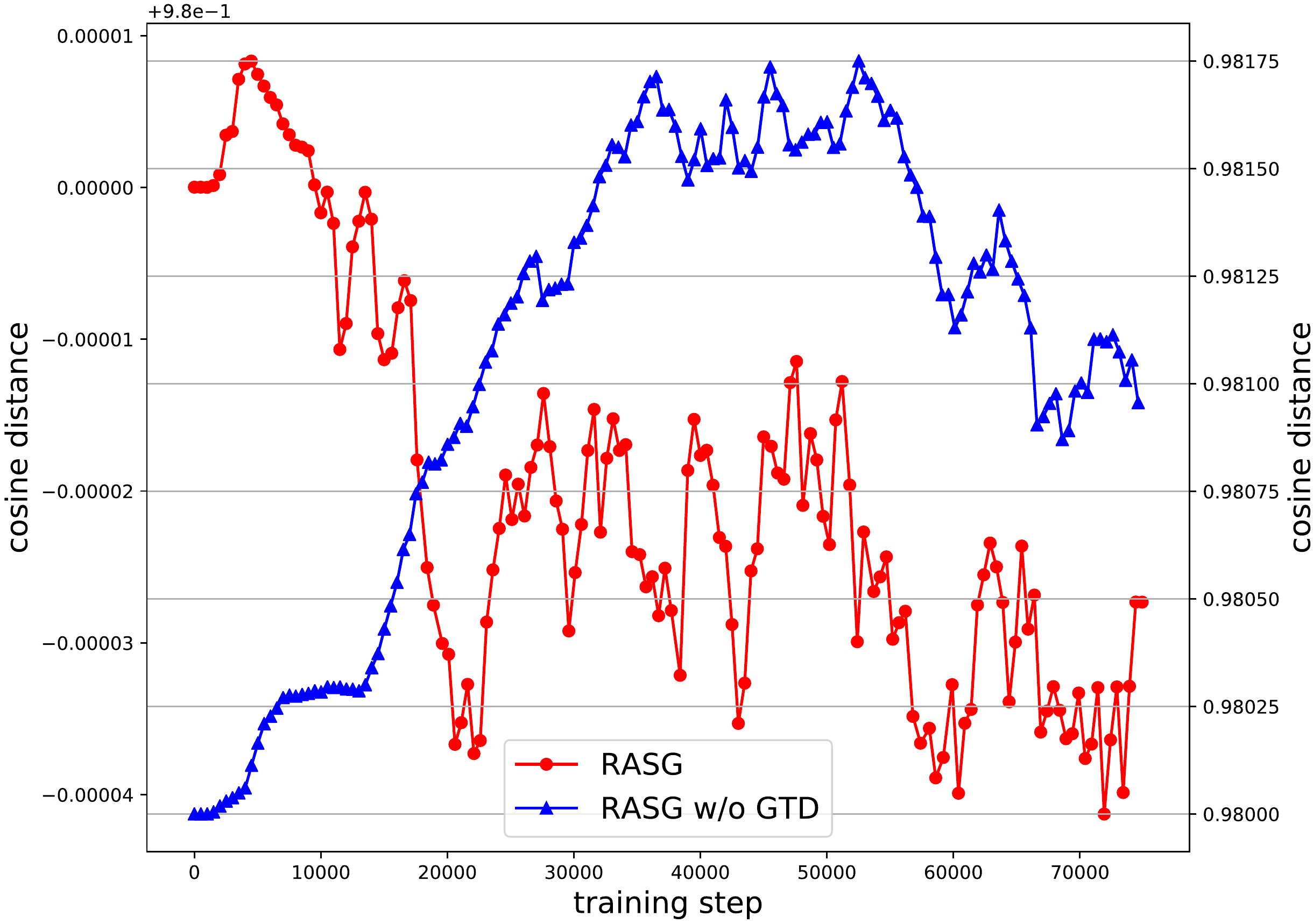}
    }
  \subfigure[]{ 
    \includegraphics[width=0.45\linewidth]{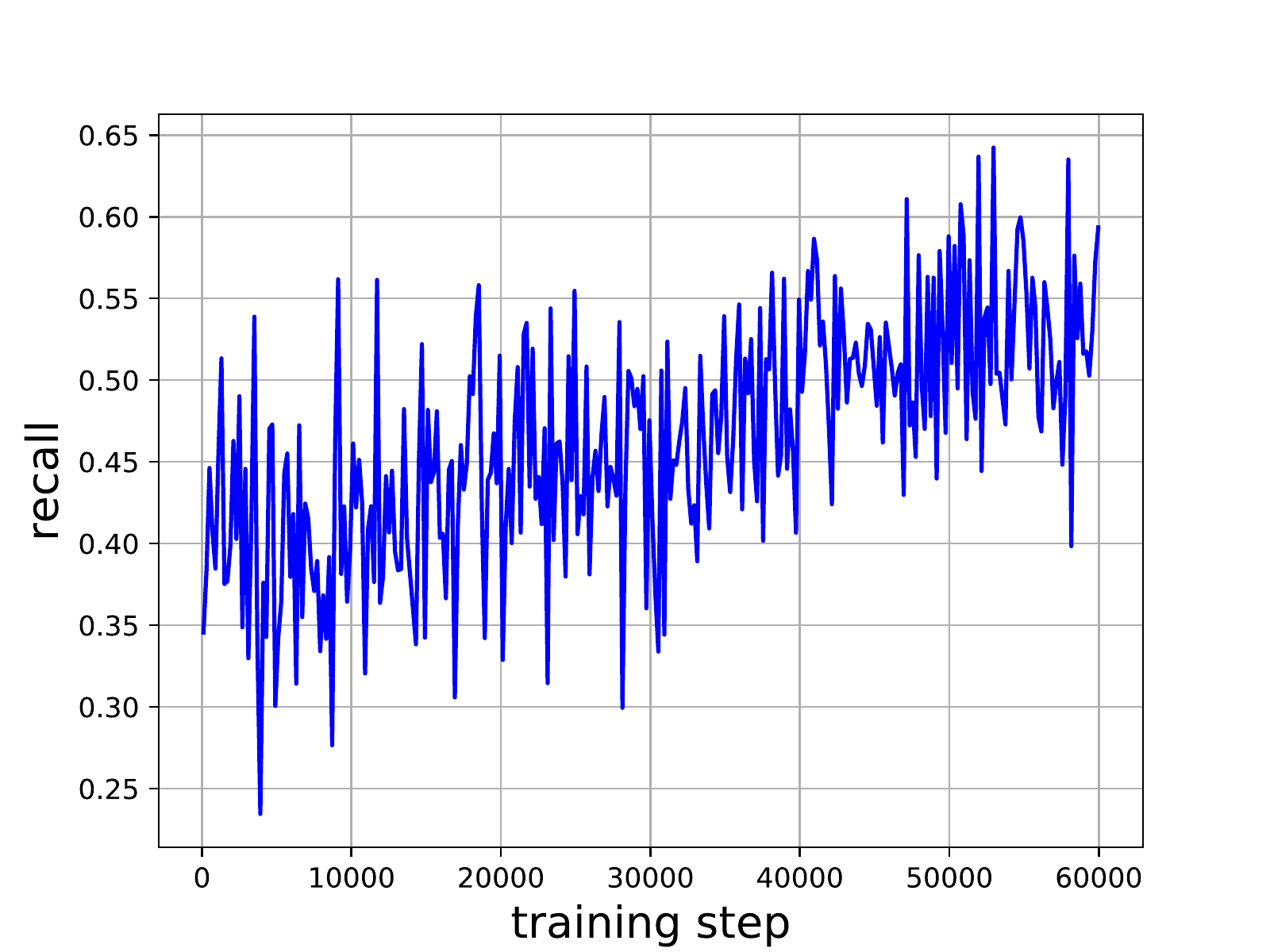}
    }
  \caption{(a) Cosine distance between decoding attention and reader attention. (b) Recall score of denoising module.} 
  \label{fig:subfig} 
\end{figure}

In this section, we turn to research question \textbf{RQ4}.
The main purpose of employing goal tracker is to help the summary generator utilize the reader focused aspect.
Intuitively, we want to know whether the summary generator follows the goal set by the goal tracker.
Therefore, we calculate the cosine distance between decoder attention $\nu_{T^{\hat{Y}}} \in \mathbb{R}^{T^d}$ (in Equation~\ref{equ:current_attention}) and reader attention $\epsilon \in \mathbb{R}^{T^d}$ (in Equation~\ref{equ:word-sim-sm}).
In Figure~\ref{fig:subfig}, we compare the cosine distance between the ablation model RASG w/o GTD and RASG.
RASG observes a decrease of cosine distance and conversely, the RASG w/o GTD observes an increment of cosine distance.
The fact that RASG can narrow the cosine distance proves that goal tracker and discriminator can lead the generator to follow the reader focused aspect.

\subsection{Human evaluation}

We ask three highly-educated Ph.D. students to rate 100 generated summaries of different models according to consistency and fluency.
These annotators are all native speakers.
The rating score ranges from 1 to 3 and 3 is the best.
We take the average score of all summaries as the final score of each model, as shown in Table~\ref{tab:comp_human_baslines}.
It can be seen that RASG outperforms other baseline models in both sentence fluency and consistency by a large margin.
We calculate the kappa statistics in terms of fluency and consistency, and the score is 0.33 and 0.29 respectively.
To prove the significance of the above results, we also do the paired student t-test between our model and CGU model (row with shaded background), the p-value are 0.0017 and 0.0012 for fluency and consistency respectively.

\begin{table}[t]
\centering
\small
\caption{Consistency and fluency comparison by human evaluation.}
\begin{tabular}{@{}lcc cc@{}}
\toprule
& \multicolumn{2}{c}{Fluency} & \multicolumn{2}{c}{Consistency} \\ \cline{2-5} 
& mean & variance  & mean & variance \\
\midrule
S2S & 2.17 & 0.24 & 1.98 & 0.28 \\
\cbkgrnd CGU & \cbkgrnd 2.20 & \cbkgrnd 0.26 & \cbkgrnd 2.08 & \cbkgrnd 0.29 \\
RASG & 2.65\dubbelop & 0.21 & 2.48\dubbelop & 0.26 \\
\bottomrule
\end{tabular}
\label{tab:comp_human_baslines}
\end{table}

\subsection{Case analysis}

\begin{figure}[!t]
    \centering
    \includegraphics[width=\linewidth]{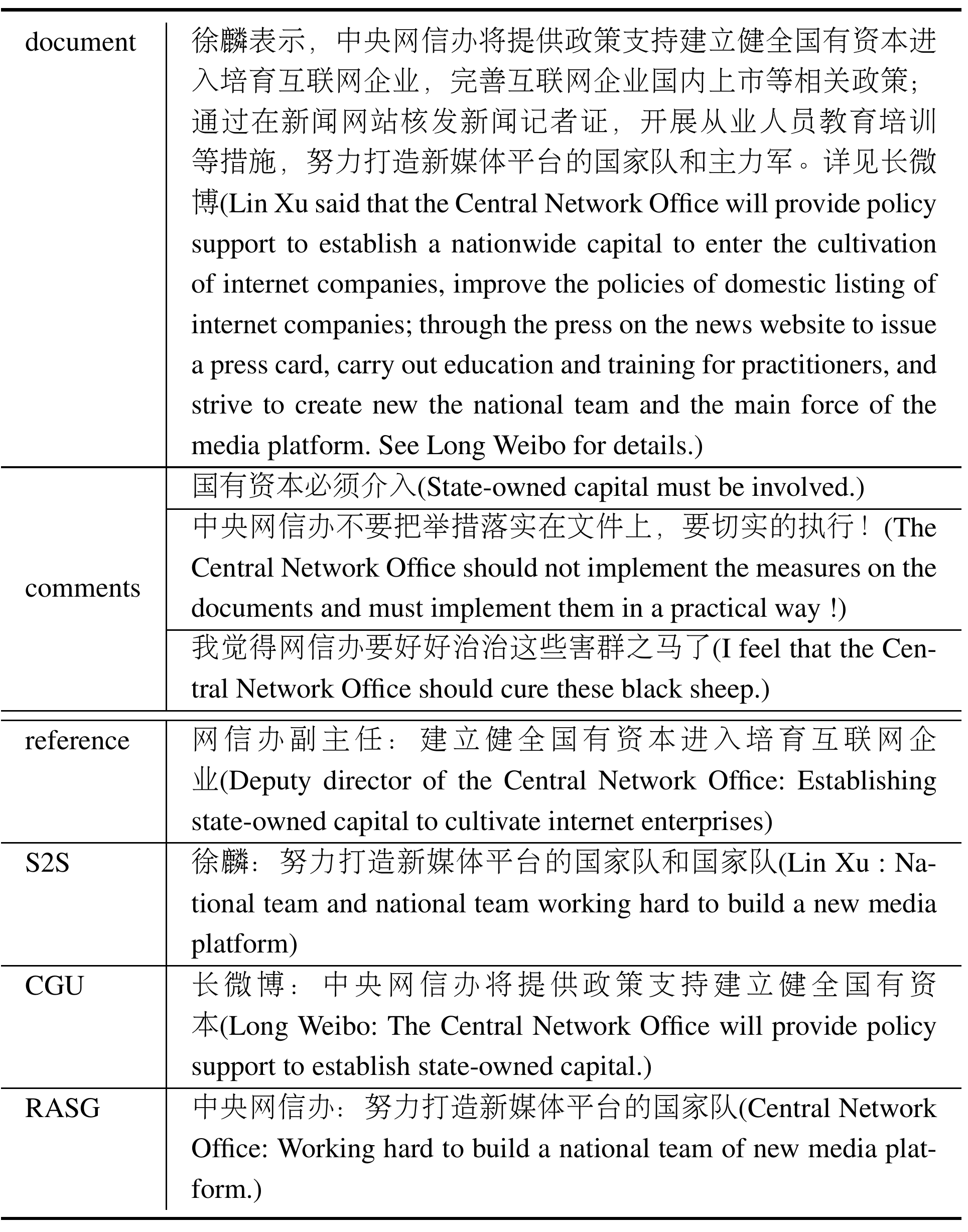}
    \caption{Examples of the generated summary by RASG and other models.}
    \label{tab:case}
\end{figure}

Figure~\ref{tab:case} shows a document and its corresponding summaries generated by different methods.
We can observe that S2S does generate fluent summary. 
However, the generated aspect is contradictory to the focused aspect of reader or ground truth summary.
Meanwhile, RASG overcomes this shortcoming by using goal vector and gap content given by goal tracker and supervisor at training stage, and produces the summary that is not only fluent but also consistent with main aspect of document.

\section{Conclusion}

In this paper, we propose a new framework named \emph{reader-aware summary generator} (RASG) which aims to generate summaries for document from social media incorporating the reader comments.
In order to capture the reader focused aspect, we design a reader attention component with a denoising module to capture the alignment between comments and document.
We employ a supervisor to measure the semantic gap between generated summary and reader focused aspect.
A goal tracker uses the information of semantic gap and the feature extracted by the discriminator to produce a goal vector to guide the summary generator.
In our experiments, we have demonstrated the effectiveness of RASG and have found significant improvements over state-of-the-art baselines in terms of ROUGE and human evaluations. 
Moreover, we have verified the effectiveness of each module in RASG for improving the summarization performance.

\section{Acknowledgments}
We would like to thank the anonymous reviewers for their constructive comments. 
We would also like to thank Zhujun Zhang, Sicong Jiang for their helps on this project. 
This work was supported by the National Key Research and Development Program of China (No. 2017YFC0804001), the National Science Foundation of China (NSFC No. 61876196, No. 61672058), Alibaba Innovative Research (AIR) Fund. 
Rui Yan was sponsored by CCF-Tencent Open Research Fund and Microsoft Research Asia (MSRA) Collaborative Research Program.

\bibliography{aaai19}
\bibliographystyle{aaai}

\end{document}